%% file: acl2019.tex
\documentclass[11pt,a4paper]{article}
\usepackage[hyperref]{acl2019}
\usepackage{times}
\usepackage{latexsym}

\usepackage{url}

\usepackage{graphicx}
\usepackage{amssymb}
\usepackage{subfig}
\usepackage{multirow}
\usepackage{xcolor}

\usepackage{xspace}
\usepackage{amsmath}
\usepackage{enumerate}

\aclfinalcopy 

\include{macros}

\title{Exploiting Explicit Paths for Multi-hop Reading Comprehension}

\author{Souvik Kundu$^{\dagger}\thanks{~~Work performed while doing an internship at the Allen Institute for Artificial Intelligence.}$ \and Tushar Khot$^{\ddagger}$ \and Ashish Sabharwal$^{\ddagger}$ \and Peter Clark$^{\ddagger}$ \\
  $^{\dagger}$Department of Computer Science, National University of Singapore \\
  $^{\ddagger}$Allen Institute for Artificial Intelligence, Seattle, WA, U.S.A. \\
  {\tt souvik@comp.nus.edu.sg, \{tushark,ashishs,peterc\}@allenai.org} 
  }

\date{}

\begin{document}
\maketitle

\begin{abstract}
We propose a novel, path-based reasoning approach for the multi-hop reading comprehension task where a system needs to combine facts from multiple passages to answer a question. Although inspired by multi-hop reasoning over knowledge graphs, our proposed approach operates directly over unstructured text. It generates potential paths through passages and scores them without any direct path supervision. The proposed model, named \modelname, attempts to extract implicit relations from text through entity pair representations, and compose them to encode each path. To capture additional context, \modelname also composes the passage representations along each path to compute a passage-based representation. Unlike previous approaches, our model is then able to explain its reasoning via these explicit paths through the passages. We show that our approach outperforms prior models on the multi-hop Wikihop dataset, and also can be generalized to apply to the OpenBookQA dataset, matching state-of-the-art performance.
\end{abstract}

\section{Introduction}
\label{sec:intro}

Many reading comprehension (RC) datasets~\cite{Rajpurkar2016-squad,Trischler2017-rc-newsqa,joshi-EtAl:2017:Trivia-qa} have been proposed recently to evaluate a system's ability to answer a question from a given text passage. However, most of the questions in these datasets can be answered by using only a single sentence or passage. 
As a result, systems designed for these tasks may not be able to compose knowledge from multiple sentences or passages, a key aspect of natural language understanding. To remedy this, new datasets~\cite{babi-Weston-15,wikihop,MultiRCKhashabi2018,openbookqa} have been proposed, requiring a system to combine information from multiple sentences in order to arrive at the answer, referred to as \emph{multi-hop reasoning}. 

\begin{figure}[t]
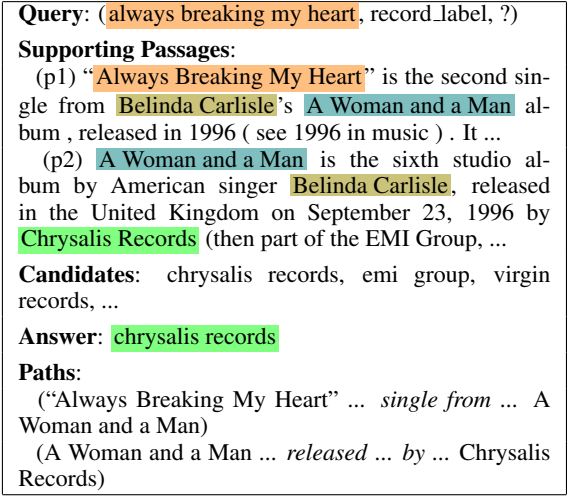

\small
\centering
\begin{tabular}{|p{7.0cm}|}
\hline
\textbf{Query}: (\entA{always breaking my heart}, record\_label, ?) \\
\T \textbf{Supporting Passages}: \\
  \hspace{1ex} (p1) ``\entA{Always Breaking My Heart}'' is the second single from \entB{Belinda Carlisle}'s \entC{A Woman and a Man} album , released in 1996 ( see 1996 in music ) . It ... \\
  \hspace{1ex} (p2) \entC{A Woman and a Man} is the sixth studio album by American singer \entB{Belinda Carlisle}, released in the United Kingdom on September 23, 1996 by \entD{Chrysalis Records} (then part of the EMI Group, ... \\
\T \textbf{Candidates}: chrysalis records, emi group, virgin records, ... \\
\T \textbf{Answer}: \entD{chrysalis records} \\
\T \textbf{Paths}: \\
  \hspace{1ex} (``Always Breaking My Heart'' ... \textit{single from} ... A Woman and a Man) \\
  \hspace{1ex} (A Woman and a Man ... \textit{released} ... \textit{by} ... Chrysalis Records)  \\ \hline

\end{tabular}
\caption{Example illustrating our proposed path extraction and reasoning approach.}
\label{tab:ex1_path_ext}
\end{figure}

Multi-hop reasoning has been studied for question answering (QA) over structured knowledge graphs~\cite{pra,kbpath,Das2017ChainsOR}. Many of the successful models explicitly identify paths in the knowledge graph that led to the answer. A strength of these models is high interpretability, arising from explicit path-based reasoning over the underlying graph structure. However, they cannot be directly applied to QA in the absence of such structure.

Consequently, most multi-hop RC models over unstructured text~\cite{GAR,hu2017reinforced} extend standard attention-based models from RC by iteratively updating the attention to indirectly ``hop'' over different parts of the text. Recently, graph-based models~\cite{mhqa_grn,gcn_entity} have been proposed for the WikiHop dataset~\cite{wikihop}. Nevertheless, these models still only implicitly combine knowledge from all passages, and are therefore unable to provide explicit reasoning paths.

We propose an approach\footnote{The source code is available at \url{https://github.com/allenai/PathNet}} for multiple choice RC that explicitly extracts potential paths from text (without direct path supervision) and encodes the knowledge captured by each path. Figure \ref{tab:ex1_path_ext} shows how to apply this approach to an example in the \wikihop dataset. It shows two sample paths connecting an entity in the question ({\em Always Breaking My Heart}) to a candidate answer ({\em Chrysalis Records}) through a singer ({\em Belinda Carlisle}) and an album ({\em A Woman and a Man}).

To encode the path, our model, named \modelname, first aims to extract implicit (latent) relations between entity pairs in a passage based on their contextual representations.
For example, it aims to extract the implicit \textit{single from} relation between the song and the name of the album in the first passage. Similarly, it extracts the \textit{released by} relation between the album and the record label in the second passage. It learns to compose the extracted implicit relations such that they map to the main relation in the query, in this case \textit{record\_label}. In essence, the motivation is to learn to extract implicit relations from text and to identify their valid compositions, such as: (x, \textit{single from}, y), (y, \textit{released by}, z) $\rightarrow$  (x, \textit{record\_label}, z). Due to the absence of direct supervision on these relations, \modelname does not explicitly extract these relations. However, our qualitative analysis on a sampled set of instances from \wikihop development set shows that the top scoring paths in 78\% of the correctly answered questions have implied relations in the text that could be composed to derive the query relations.

In addition, \modelname also learns to compose aggregated passage representations in a path to capture more global information: encoding(p1),  encoding(p2) $\rightarrow$ (x, \textit{record\_label}, z). This passage-based representation is especially useful in domains such as science question answering where the lack of easily identifiable entities limits the effectiveness of the entity-based path representation. While this passage-based representation is less interpretable than the entity-based path representation, it still identifies the two passages used to select the answer, compared to a spread out attention over all documents produced by previous graph-based approaches.

We make three main contributions:

(1) A novel path-based reasoning approach for multi-hop QA over text that produces explanations in the form of explicit paths;
(2) A model, \modelname, which aims to extract implicit relations from text and compose them; and
(3) Outperforming prior models on the target \wikihop dataset\footnote{Other systems, such as by \citet{cfc}, have recently appeared on the \wikihop leaderboard (\url{http://qangaroo.cs.ucl.ac.uk/leaderboard.html}).} and generalizing to the open-domain science QA dataset, OpenBookQA, with performance comparable to prior models.

\section{Related Work}
\label{sec:related_work}
We summarize related work in QA over text, semi-structured knowledge, and knowledge graphs. 

\textbf{Multi-hop RC.}
Recent datasets such as bAbI~\cite{babi-Weston-15}, Multi-RC~\cite{MultiRCKhashabi2018}, \wikihop~\cite{wikihop}, and OpenBookQA~\cite{openbookqa} have encouraged research in multi-hop QA over text. The resulting multi-hop models can be categorized into state-based and graph-based reasoning models. State-based reasoning models~\cite{GAR,shen2017reasonet,hu2017reinforced} are closer to a standard attention-based RC model with an additional ``state'' representation that is iteratively updated. The changing state representation results in the model focusing on different parts of the passage during each iteration, allowing it to combine information from different parts of the passage. Graph-based reasoning models~\cite{corefgru,gcn_entity,mhqa_grn}, on the other hand, create graphs over entities within the passages and update entity representations via recurrent or convolutional networks. In contrast, our approach explicitly identifies paths connecting entities in the question to the answer choices.

\textbf{Semi-structured QA.}
Our model is closer to Integer Linear Programming (ILP) based methods~\cite{tableilp2016,Khot2017AnsweringCQ,semanticilp2018aaai}, which define an ILP program to find optimal {\em support graphs} for connecting the question to the choices through a semi-structured knowledge representation. 
However, these models require a manually authored and tuned ILP program, and need to convert text into a semi-structured representation---a process that is often noisy (such as using Open IE tuples~\cite{Khot2017AnsweringCQ}, SRL frames~\cite{semanticilp2018aaai}). Our model, on the other hand, is trained end-to-end, and discover relevant relational structure from text. Instead of an ILP program, \namecite{Jansen2017FramingQA} train a latent ranking perceptron using features from aggregated syntactic structures from multiple sentences. However, their system operates at the detailed (and often noisy) level of dependency graphs, whereas we identify entities and let the model learn implicit relations and their compositions. 

\textbf{Knowledge Graph QA.}
QA datasets on knowledge graphs such as Freebase~\cite{bollacker2008freebase}, require systems to map queries to a single relation~\cite{simplequestions}, a path~\cite{kbpath}, or complex structured queries~\cite{webquestions} over these graphs. While early models~\cite{pra,sfe} focused on creating path-based features, recent neural models~\cite{kbpath,Das2017ChainsOR,Toutanova2016CompositionalLO} encode the entities and relations along a path and compose them using recurrent networks. Importantly, the input knowledge graphs have entities and relations that are shared across all training and test examples, which the model can exploit during learning (e.g., via learned entity and relation embeddings). When reasoning with text, our model must learn these representations purely based on their local context.
\section{Approach Overview}
\label{sec:prob_def}
We focus on the multiple-choice RC setting: given a question and a set of passages, the task is to find the correct answer among a predefined set of candidates. 
The proposed approach can be applied to $m$-hop reasoning, as discussed briefly in the corresponding sections for path extraction, encoding, and scoring. Since our target datasets primarily need 2-hop reasoning\footnote{We found that most \wikihop questions can be answered with 2 hops and OpenBookQA also targets 2-hop questions.} and the potential of semantic drift with increased number of hops~\cite{fried2015higher,Khashabi2019OnTC}, we focus on and assess the case of 2-hop paths ($m=2$). 
 As discussed later (see Footnote~\ref{footnote:scaling}), our path-extraction step scales exponentially with $m$. Using $m=2$ keeps this step tractable, while still covering almost all examples in our target datasets.

In \wikihop, a question $\ques$ is given in the form of a tuple $(h_e, r, ?)$, where $h_e$ represents the head entity and $r$ represents the relation between $h_e$ and the unknown tail entity. The task is to select the unknown tail entity from a given set of candidates $\{c_1, c_2, \ldots c_N\}$, by reasoning over supporting passages $\psgs = {\psg_1, \ldots, \psg_M}$. 
To perform multi-hop reasoning, we extract multiple paths $\pths$ (cf.~Section~\ref{sec:path_extract}) connecting $h_e$ to each $c_k$ from the supporting passages $\psgs$. The $j$-th 2-hop path for candidate $c_k$ is denoted $\pth_{kj}$,
where $\pth_{kj}=h_e \rightarrow e_1 \rightarrow c_k$, and $e_1$ is referred to as the \emph{intermediate entity}.

In OpenBookQA, different from \wikihop, the questions and candidate answer choices are plain text sentences. To construct paths, we extract all head entities from the question and tail entities from candidate answer choices, considering all noun phrases and named entities as entities. This often results in many 2-hop paths connecting a question to a candidate answer choice via the same intermediate entity. With $\{h_{e_1}, h_{e_2}, \ldots\}$ representing the list of head entities from a question, and $\{c_{k_1}, c_{k_2}, \ldots\}$ the list of tail entities from candidate $c_k$, the $j$-th path connecting $c_{k_\alpha}$ to $h_{e_\beta}$ can be represented as: $\pth_{kj}^{\alpha,\beta} = h_{e_\alpha} \rightarrow e_1 \rightarrow c_{k_\beta}$. For simplicity, we omit the notations $\alpha$ and $\beta$ from path representation. 

Next, the extracted paths are encoded and scored (cf.~Section \ref{sec:model}). Following, the normalized path scores are summed for each candidate to give a probability distribution over the candidate answer choices.

\section{Path Extraction}
\label{sec:path_extract}

The first step in our approach is extracting paths from text passages. Consider the example in Figure \ref{tab:ex1_path_ext}.
Path extraction proceeds as follows: 

(a) We find a passage $\psg_1$ that contains a head entity $h_e$ from the question $\ques$. In our example, we would identify the first supporting passage that contains {\em always breaking my heart}. 

(b) We then find all named entities and noun phrases that appear in the same sentence as $h_e$ or in the subsequent sentence. Here, we would collect {\em Belinda Carlisle}, {\em A Woman and a Man}, and {\em album} as potential intermediate entity $e_1$. 

(c) Next, we find a passage $\psg_2$ that contains the
potential intermediate entity identified above. For clarity, we refer to the occurrence of $e_1$ in $\psg_2$ as $e_1\prime$. By design, ($h_e, e_1$) and ($e_1\prime, c_k$) are located in different passages. For instance, we find the second passage that contains both {\em Belinda Carlisle} and {\em A Woman and a Man}.

(d) Finally, we check whether $\psg_2$ contains any of the candidate answer choices. For instance, $\psg_2$ contains {\em chrysalis records} and {\em emi group}.

The resulting extracted paths can be summarized as a set of entity sequences. In this case, for the candidate answer {\em chrysalis records}, we obtain a set of two paths: {\em (always breaking my heart $\rightarrow$ Belinda Carlisle $\rightarrow$ chrysalis records), (always breaking my heart $\rightarrow$ A Man and a Woman $\rightarrow$ chrysalis records)}.
Similarly, we can collect paths for the other candidate, {\em emi group}.

Notably, our path extraction method can be easily extended for more hops.
Specifically, for $m$-hop reasoning, steps (b) and (c) are repeated $(m - 1)$ times, where the intermediate entity from step (c) becomes the head entity for the subsequent step (b). For larger values of $m$, maintaining tractability of this approach  would require optimizing the complexity of identifying the passages containing an entity (steps (a) and (c)) and limiting the number of neighboring entities considered (step (b)).\footnote{\label{footnote:scaling}If the search step takes no more than $s$ steps and identifies a fixed number $k$ of passages, and we select up to $e$ neighboring entities, our approach would have a time complexity of O$\left((ke)^{m-1}s^{m}\right)$ for enumerating $m$-hop paths.}

For one hop reasoning, i.e., when a single passage is sufficient to answer a question, we construct the path with $e_1$ as {\em null}. In this case, both $h_e$ and $c_k$ are found in a single passage.
In this way, for a task requiring more hops, one only need to guess the maximum number of hops. If some questions in that task require less hops, our proposed approach can easily handle that by assigning the intermediate entity to \emph{null}. For instance, in this work, our approach can handle 1-hop reasoning although it is developed for 2-hop.

\section{\model Model}
\label{sec:model}

\begin{figure}[t]
\centering
\includegraphics[width=0.38\textwidth]{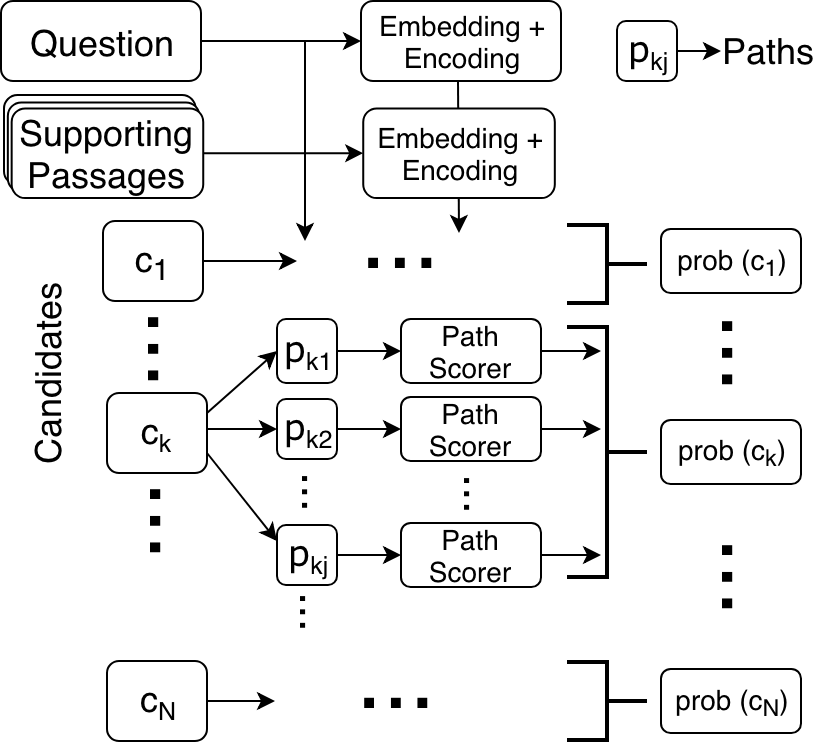}
\caption{Architecture of the proposed model.}
\label{fig:fullmodel_diagram}
\end{figure}

Once we have all potential paths, we score them using the proposed model, named PathNet, whose overview is depicted in Figure~\ref{fig:fullmodel_diagram}. The key component is the path-scorer module that computes the score for each path $\pth_{kj}$.
We normalize these scores across all paths, and compute the probability of a candidate $c_k$ being the correct answer by summing the normalized scores of the paths associated with $c_k$:
\begin{equation}
    \textnormal{prob}(c_k) = \sum_j \textnormal{score}(\pth_{kj}).
\end{equation}

\begin{figure*}[t]
\centering
\includegraphics[width=0.9\textwidth]{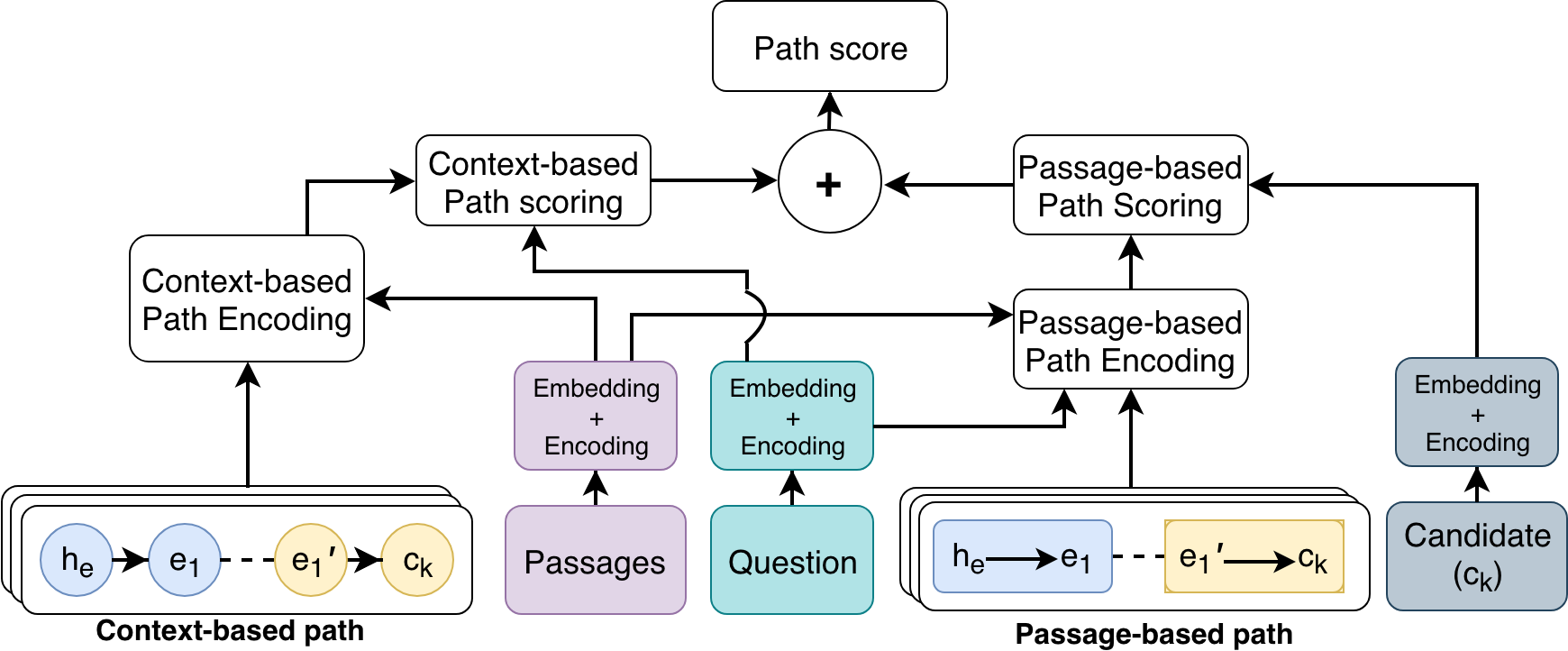}
\caption{Architecture of the path scoring module, shown here for 2-hop paths.}
\label{fig:path_enc_diagram}
\end{figure*}

Next, we describe three main model components, operating on the following inputs: question $\ques$, passages $\psg_1$ and $\psg_2$, candidate $c_k$, and the locations of $h_e$, $e_1$, $e_1^\prime$, $c_k$ in these passages: 
(1) Embedding and Encoding (\S~\ref{subsec:enc})
(2) Path Encoding (\S~\ref{subsec:path_enc})
(3) Path Scoring (\S~\ref{subsec:path_score}).
In Figure \ref{fig:path_enc_diagram}, we present the model architecture for these three components used for scoring the paths.

\subsection{Embedding and Encoding}
\label{subsec:enc}
We start by describing how we embed and contextually encode all pieces of text: question, supporting passages, and candidate answer choices. 
For word embedding, we use pretrained 300 dimensional vectors from GloVe \cite{pennington2014glove}, randomly initializing vectors for out of vocabulary (OOV) words. 
For contextual encoding, we use bi-directional LSTM (BiLSTM) \cite{LSTM}.

Let $T$, $U$, and $V$ represent the number of tokens in the $p$-th supporting passage, question, and $k$-th answer candidate, respectively.
The final encoded representation for the $p$-th supporting passage can be obtained by stacking these vectors into $\psgsmat_p \in \mathbb{R} ^{T \times H}$, where $H$ is the number of hidden units for the BiLSTMs.
The sequence level encoding for the question, $\mathbf{Q} \in \mathbb{R} ^{U \times H}$, and for the $k$-th candidate answer, $\mathbf{C}_k \in \mathbb{R} ^{V \times H}$, are obtained similarly.
We use row vector representation (e.g., $\mathbb{R}^{1 \times H}$) for all vectors in this paper. 

\subsection{Path Encoding}
\label{subsec:path_enc}
After extracting the paths as discussed in Section \ref{sec:path_extract}, they are encoded using an end-to-end neural network.
This path encoder consists of two components: context-based and passage-based.
\subsubsection{Context-based Path Encoding}
\label{sec:ctx_path_enc}
This component aims to implicitly encode the relation between $h_e$ and $e_1$, and between $e_1\prime$ and $c_k$.
These implicit relation representations are them composed together to encode a path representation for $h_e \rightarrow e_1 \ldots e_1\prime \rightarrow c_k$. 

First, we extract the contextual representations for each of $h_e$, $e_1$, $e_1\prime$, and $c_k$. Based on the \emph{locations} of these entities in the corresponding passages, we extract the boundary vectors from the passage encoding representation. For instance, if $h_e$ appears in the $p$-th supporting passage from token $i_1$ to $i_2$ ($i_1 \leq i_2$), then the contextual encoding of $h_e$, $\locvec_{h_e} \in \mathbb{R} ^{2H}$ is taken to be:
$\locvec_{h_e} = \mathbf{s}_{p_1, i_1} ~||~ \mathbf{s}_{p_1, i_2}$, where $||$ denotes the concatenation operation. 
If $h_e$ appears in multiple locations within the passage, we use the mean vector representation across all of these locations. 
The location encoding vectors $\locvec_{e_1}$, $\locvec_{e_1\prime}$, and $\locvec_{c_k}$ are obtained similarly.

Next, we extract the implicit relation between $h_e$ and $e_1$ as $\mathbf{r}_{h_e,e_1} \in \mathbb{R} ^{H}$, using a feed forward layer: 
\begin{equation}
    \mathbf{r}_{h_e, e_1} = \textnormal{FFL}(\locvec_{h_e} , \locvec_{e_1} ) ~~,
\end{equation}
where FFL is defined as:
\begin{equation}
\textnormal{FFL}(\mathbf{a}, \mathbf{b}) = \textnormal{tanh}(\mathbf{a} \mathbf{W}_a ~+~ \mathbf{b} \mathbf{W}_b)  ~~.
\end{equation}
Here $\mathbf{a} \in \mathbb{R} ^{H^{\prime}}$ and $\mathbf{b} \in \mathbb{R} ^{H^{\prime \prime}}$ are input vectors, and
$\mathbf{W}_a \in \mathbb{R}^{H^{\prime} \times H}$ and $\mathbf{W}_b \in \mathbb{R}^{H^{\prime \prime} \times H}$ are trainable weight matrices. The bias vectors are not shown here for simplicity.
Similarly, we compute the implicit relation between $e_1\prime$ and $c_k$ as $\mathbf{r}_{e_1\prime, c_k} \in \mathbb{R} ^{H}$, using their location encoding vectors $\locvec_{e_1\prime}$ and $\locvec_{c_k}$.

Finally, we \emph{compose} all implicit relation vectors along the path to obtain a context-based path representation $\mathbf{x}_{\textnormal{ctx}} \in \mathbb{R} ^{H}$ given by:
\begin{equation}
\mathbf{x}_{\textnormal{ctx}} = \textnormal{comp}(\mathbf{r}_{h_e,e_1} ~,~
\mathbf{r}_{e_1\prime,c_k})
\label{eqn:compsn}
\end{equation}

For fixed length paths, we can use a feed forward network as the composition function. E.g., for 2-hop paths, we use $\textnormal{FFL}(\mathbf{r}_{h_e,e_1} ~,~
\mathbf{r}_{e_1\prime,c_k})$. For variable length paths, we can use recurrent composition networks such as LSTM, GRU. We compare these composition functions in Section~\ref{subsec:path_comp_eval}. 
\subsubsection{Passage-based Path Encoding}
In this encoder, we use \emph{entire passages} to compute the path representation. As before, suppose ($h_e, e_1$) and ($e_1\prime, c_k$) appear in supporting passages $p_1$ and $p_2$, respectively. 
We encode each of $p_1$ and $p_2$ into a single vector based on passage-question interaction.
As discussed below, we first compute a question-weighted representation for passage tokens and then aggregate it across the passage. 

\paragraph{Question-Weighted Passage Representation:}
For the $p$-th passage, we first compute the attention matrix $\mathbf{A} \in \mathbb{R} ^{T \times U}$, capturing the similarity between the passage and question words. Then, we calculate a question-aware passage representation $\mathbf{S}^{q_1}_p \in \mathbb{R} ^ {T \times H}$, where $\mathbf{S}^{q_1}_p = \mathbf{A}\mathbf{Q}$. 
Similarly, a passage-aware question representation, $\mathbf{Q}_p \in \mathbb{R} ^ {U \times H}$, is computed, where $\mathbf{Q}_p = \mathbf{A}^\top \mathbf{S}_p$. 

Further, we compute another passage representation $\mathbf{S}^{q_2}_p = \mathbf{A} \mathbf{Q}_p \in \mathbb{R}^{T \times H}$. Intuitively, $\mathbf{S}^{q_1}_p$ captures important passage words based on the question, whereas $\mathbf{S}^{q_2}_p$ is another passage representation which focuses on the interaction with passage-relevant question words.
The idea of encoding a passage after interacting with the question multiple times is inspired from the Gated Attention Reader model~\cite{GAR}.

\paragraph{Aggregate Passage Representation:}
To derive a single passage vector, we first concatenate the two passage representations for each token, obtaining $\mathbf{S}^q_p = \mathbf{S}^{q_1}_p ~||~ \mathbf{S}^{q_2}_p \in \mathbb{R}^{T \times 2H}$. We then use an attentive pooling mechanism for aggregating the token representations. The aggregated vector $\tilde{\mathbf{s}}_{p} \in \mathbb{R} ^{2H}$ for the $p$-th passage is obtained as:
\begin{eqnarray}
\label{eq:attn_pooling}
a^{p}_t  \propto  \textnormal{exp}(\mathbf{s}^{q}_{p,t}\mathbf{w}^\top) ;~~ \tilde{\mathbf{s}}_{p}  =  \mathbf{a}^{p} \mathbf{S}^{q}_{p}
\end{eqnarray}
where $\mathbf{w} \in \mathbb{R} ^{2H}$ is a learned vector. In this way, we obtain the aggregated vector representations for both supporting passages $p_1$ and $p_2$ as $\tilde{\mathbf{s}}_{p_1} \in \mathbb{R} ^{2H}$ and $\tilde{\mathbf{s}}_{p_2} \in \mathbb{R} ^{2H}$, respectively.

\paragraph{Composition:}
We compose the aggregated passage vectors to obtain the passage-based path representation $\mathbf{x}_{\textnormal{psg}} \in \mathbb{R} ^{H}$ similar to Equation~\ref{eqn:compsn}: 
\begin{equation}
    \mathbf{x}_{\textnormal{psg}} = \textnormal{comp}(\tilde{\mathbf{s}}_{p_1} ~,~ \tilde{\mathbf{s}}_{p_2} )
\end{equation}{}

Similar to the composition function in context-based path encoding, this composition function can be a feed-forward network for fixed length or recurrent networks for variable length paths.

\subsection{Path Scoring}
\label{subsec:path_score}
Encoded paths are scored from two perspectives.

\paragraph{Context-based Path Scoring:}
We score context-based paths based on their interaction with the question encoding.
First, we aggregate the question into a single vector. 
We take the first and last hidden state representations from the question encoding $\mathbf{Q}$ to obtain an aggregated question vector representation.

The aggregated question vector $\tilde{\mathbf{q}} \in \mathbb{R} ^{H}$ is 
\begin{align}
\tilde{\mathbf{q}} ~=~ (\mathbf{q}_0 ~||~ \mathbf{q}_U) ~ \mathbf{W}_q  ~~,
\end{align}
where $\mathbf{W}_q \in \mathbb{R} ^{2H \times H}$ is a learnable weight matrix. The combined representation $\mathbf{y}_{x_{\textnormal{ctx}}, q} \in \mathbb{R} ^{H}$ of the question and a context-based path is computed as: $\mathbf{y}_{x_{\textnormal{ctx}},q} = \textnormal{FFL}(\mathbf{x}_{\textnormal{ctx}} ~,~
\tilde{\mathbf{q}} )$
Finally, we derive scores for context-based paths: 
\begin{equation}
z_{\textnormal{ctx}} = \mathbf{y}_{x_{\textnormal{ctx}},q} \mathbf{w}^{\top}_{\textnormal{ctx}}  ~~,
\end{equation}
where $\mathbf{w}_{\textnormal{ctx}} \in \mathbb{R} ^{H}$ is a trainable vector.
\paragraph{Passage-based Path Scoring:}
We also score paths based on the interaction between the passage-based path encoding vector and the candidate encoding. In this case, only candidate encoding is used since passage-based path encoding already uses the question representation.
We aggregate the representation $\mathbf{C}_k$ for candidate $c_k$ into a single vector $\tilde{\mathbf{c}}_k \in \mathbb{R} ^{H}$ by applying an attentive pooling operation similar to Equation~\ref{eq:attn_pooling}. The score for passage-based path is then computed as follows: 
\begin{equation}
    z_{\textnormal{psg}} = \tilde{\mathbf{c}}_k  ~ \mathbf{x}^{\top}_{\textnormal{psg}}
\end{equation}

Finally, the unnormalized score for path $\pth_{kj}$ is:
\begin{equation}
z = z_{\textnormal{ctx}} ~+~ z_{\textnormal{psg}} 
\end{equation}
and its normalized version, $score(\pth_{kj})$, is calculated by applying the softmax operation over all the paths and candidate answers.


\section{Experiments}
\label{sec:experiments}
We start by describing the experimental setup, and then present results and an analysis of our model.

\subsection{Setup}
We consider the standard (unmasked) version of the recently proposed \wikihop dataset \cite{wikihop}.
\wikihop is a large scale multi-hop QA dataset consisting of about 51K questions (5129 Dev, 2451 Test). 
Each question is associated with an average of 13.7 supporting Wikipedia passages, each with 36.4 tokens on average.

We also evaluate our model on OpenBookQA~\cite{openbookqa}, a very recent and challenging multi-hop QA dataset with about 6K questions (500 Dev, 500 Test), each with 4 candidate answer choices. Since OpenBookQA does not have associated passages for the questions, we retrieve  sentences from a text corpus to create single sentence passages. 

We start with a corpus of 1.5M sentences used by previous systems~\cite{Khot2017AnsweringCQ} for science QA. It is then filtered down to 590K sentences by identifying sentences about generalities and removing noise. We assume sentences that start with a plural noun are likely to capture general concepts, e.g. ``Mammals have fur'', and only consider such sentences. We also eliminate noisy and irrelevant sentences by using a few rules such as root of the parse tree must be a sentence, it must not contain proper nouns. This corpus is also provided along with our code.

Next, we need to retrieve sentences that can lead to paths between the question $q$ and an answer choice $c$. Doing so naively will only retrieve sentences that directly connect entities in $q$ to $c$, i.e., 1-hop paths. To facilitate 2-hop reasoning, we first retrieve sentences based on words in $q$, and for each retrieved sentence $s_1$, we find sentences that overlap with both $s_1$ and $c$. Each path is scored using $\textnormal{idf}(q, s_1)\cdot \textnormal{idf}(s_1, s_2)\cdot \textnormal{idf}(s_2, c)$, where $s_2$ is the second retrieved sentence and $\textnormal{idf}(w)$ is the idf score of token $w$ based on the input corpus:
\[
\textnormal{idf}(x, y)=\frac{\sum_{w \in x \cap y} \textnormal{idf}(w)}{\min(\sum_{w \in x} \textnormal{idf}(w),\sum_{w \in y} \textnormal{idf}(w))}
\]
For efficiency, we perform beam search and ignore any chain if the score drops below a threshold (0.08). Finally we take the top 100 chains and use these sentences as passages in our model.

We use Spacy\footnote{\url{https://spacy.io/api/tokenizer}} for tokenization. For word embedding, we use the 840B 300-dimensional pre-trained word vectors from GloVe and we do not update them during training. 
For simplicity, we do not use any character embedding. The number of hidden units in all LSTMs is 50 ($H=100$). 
We use dropout \cite{dropout} with probability 0.25 for every learnable layer. During training, the minibatch size is fixed at 8. We use the Adam optimizer \cite{adam} with learning rate 0.001 and clipnorm 5. We use cross entropy loss for training. This being a multiple-choice QA task, we use accuracy as the evaluation metric.

\subsection{Main Results}
\begin{table}[t]
\small
\centering
\setlength{\tabcolsep}{10pt}
\setlength{\doublerulesep}{\arrayrulewidth}
\begin{tabular}{l|cc}
\hline
\multirow{2}{*}{Model} & \multicolumn{2}{c}{Accuracy (\%)} \\ \cline{2-3} 
 & Dev & Test \\ \hline \hline
\T \citet{wikihop} & - & 42.9 \\ 
\citet{corefgru} & 56.0 & 59.3 \\ 
\citet{mhqa_grn} & 62.8 & 65.4 \\ 
\citet{gcn_entity} & 64.8 & 67.6 \\ 
\hline
\modelname & \textbf{67.4}{$^\dagger$} & \textbf{69.6}{$^\dagger$} \\ \hline
\end{tabular}
\caption{Accuracy on the \wikihop dataset. {$^\dagger$}Statistically significant \citep{wilson}}
\label{tab:comparison_results}
\end{table}
\begin{table}[t]
\small
\centering
\setlength{\tabcolsep}{10pt}
\setlength{\doublerulesep}{\arrayrulewidth}
\begin{tabular}{l|cc}
\hline
\multirow{2}{*}{Model} & \multicolumn{2}{c}{Accuracy (\%)} \\ \cline{2-3} 
 & Dev & Test \\ \hline \hline
\T KER (OMCS) & 54.4 & 52.2 \\ 
KER (WordNet) & 55.6	 & 51.4 \\ 
KER (OB + OMCS)  & 54.6 & 50.8 \\ 
KER (OB + WordNet) & 54.2 & 51.2 \\ 
KER (OB + Text) & 55.4 & 52.0 \\ 
\hline
\modelname(OB + Text) & 55.0 & \textbf{53.4} \\ \hline
\end{tabular}
\caption{Accuracy on the OpenBookQA dataset.}
\label{tab:comparison_results_obqa}
\end{table}

Table \ref{tab:comparison_results} compares our results on the \wikihop dataset with several recently proposed multi-hop QA models. We show the best results from each of the competing entries. 
\citet{wikihop} presented the results of BiDAF \cite{bidaf} on the \wikihop dataset. \citet{corefgru} incorporated coreference connections inside GRU network to capture coreference links while obtaining the contextual representation. Recently, \citet{gcn_entity} and \citet{mhqa_grn} proposed graph neural network approaches for multi-hop reading comprehension. While the high level idea is similar for these work, \citet{gcn_entity} used ELMo~\cite{elmo} for a contextual embedding, which has proven to be very useful in the recent past in many NLP tasks.

As seen in Table \ref{tab:comparison_results}, our proposed model \modelname significantly outperforms prior approaches on \wikihop.
Additionally, we benefit from \emph{interpretability}: unlike these prior methods, our model allows identifying specific entity chains that led to the predicted answer.

Table \ref{tab:comparison_results_obqa} presents results on the OpenBookQA dataset. We compare with the Knowledge Enhanced Reader (KER) model~\cite{openbookqa}. The variants reflect the source from which the model retrieves relevant knowledge: the open book (OB), WordNet subset of ConceptNet, and Open Mind Common Sense (OMCS) subset of ConceptNet, and the corpus of 590K sentences (Text). Since KER does not scale to a corpus of this size, we provided it with the combined set of sentences retrieved by our model for all the OpenBookQA questions. The model computes various cross-attentions between the question, knowledge, and answer choices, and combines these attentions to select the answer. Overall, our proposed approach marginally improved over the previous models on the OpenBookQA dataset\footnote{\namecite{Sun2018ImprovingMR} used the large OpenAI fine-tuned language model~\cite{radford2018improving} pre-trained on an additional dataset, RACE~\cite{Lai2017RACELR} to achieve a score of 55\% on this task.}. Note that, our model was designed for the closed-domain setting where all the required knowledge is provided. Yet, our model is able to generalize on the open-domain setting where the retrieved knowledge may be noisy or insufficient to answer the question.

\subsection{Effectiveness of Model Components}
\label{subsec:path_comp_eval}
Table \ref{tab:ablation_results} shows the impact of context-based and passage-based path encodings.
Performance of the model degrades when we ablate either of the two path encoding modules. 
Intuitively, in context-based path encodings, limited and more fine-grained context is considered due to the use of specific entity locations.
On the contrary, the passage-based path encoder computes the path representations considering the entire passage representations (both passages which contain the head entity and tail entity respectively). As a result, even if the intermediate entity can not be used meaningfully, the model poses the ability to form an implicit path representation. 
Passage-based path encoder is more helpful on OpenBookQA as it is often difficult to find meaningful explicit context-based paths through entity linking across passages.
Let us consider the following example taken from OpenBookQA development set where our model successfully predicted the correct answer.
\\ \\
Question: What happens when someone on top of a \textbf{bicycle} starts pushing it 's peddles in a circular motion ? \\
Answer: the \textit{bike} accelerates \\
Best Path: (\textbf{bicycle}, \underline{pedal}, \textit{bike}) \\
p1: \textbf{bicycles} require continuous circular motion on \underline{pedals} \\ 
p2: pushing on the \underline{pedals} of a \textit{bike} cause that bike to move.

In this case, the extracted path through entity linking is not meaningful as the path composition would connect bicycles to bike~
\footnote{Entities in science questions can be phrases and events (e.g., ``the bike accelerates''). Identifying and matching such entities are very challenging in case of the OpenBookQA dataset. We show that our entity-linking approach, designed for noun phrases and named entities, is still able to perform comparable to state-of-the-art methods on science question answering, despite this noisy entity matching.}. However, when the entire passages are considered, they contain sufficient information to help infer the answer.

Table \ref{tab:composition_funcs} presents the results on \wikihop development set when different composition functions are used for Equation~(\ref{eqn:compsn}). 
Recurrent networks, such as LSTM and GRU, enable the path encoder to model an arbitrary number of hops. For 2-hop paths, we found that a simple feed forward network (FFL) performs slightly better than the rest. We also considered sharing the weights (FFL shared) when obtaining the relation vectors $\mathbf{r}_{h_e, e_1}$ and $\mathbf{r}_{e_1\prime, c_k}$. Technically, the FFL model is performing the same task in both cases: extracting implicit relations and the parameters could be shared.
However, practically, the unshared weights perform better, possibly because it gives the model the freedom to handle answer candidates differently, especially allowing the model to consider the likelihood of a candidate being a valid answer to \emph{any} question, akin to a prior. 

\begin{table}[t]
\small
\centering
\setlength{\tabcolsep}{6pt}
\setlength{\doublerulesep}{\arrayrulewidth}
\begin{tabular}{l|cc}
\hline
\multirow{3}{*}{Model} & \multicolumn{2}{c}{\% Accuracy ($\Delta$)} \\ \cline{2-3} 
\T & \wikihop & OBQA \\ 
 \hline \hline
\T \modelname & \textbf{67.4}{$^\dagger$}\phantom{ (0.0)} & \textbf{55.0}{$^\dagger$}\phantom{ (0.0)} \\
~~- context-based path & 64.7 (2.7) & 54.8{$^\ast$} (0.2)\\
~~- passage-based path & 63.2 (4.2) & 46.2 (8.8) \\
\hline
\end{tabular}
\caption{Ablation results on development sets. {$^\ast$}Improvement over this is not statistically significant.}
\label{tab:ablation_results}
\end{table}
\begin{table}[t]
\small
\centering
\setlength{\tabcolsep}{10pt}
\setlength{\doublerulesep}{\arrayrulewidth}
\begin{tabular}{l|cc}
\hline
\multirow{2}{*}{Model} & \multicolumn{2}{c}{Accuracy (\%)} \\ \cline{2-3} 
\T & \wikihop & $\Delta$ \\ \hline \hline
\T FFL (\modelname) & \textbf{67.4} & - \\ 
FFL Shared & 66.7 & 0.7 \\ 
LSTM & 67.1 & 0.3 \\ 
GRU & 67.3 & 0.1 \\ \hline
\end{tabular}
\caption{Various composition functions to generate path representation ($\mathbf{x}_{\textnormal{ctx}}$) on \wikihop development set.}
\label{tab:composition_funcs}
\end{table}

\begin{table*}[t]
\small
\centering
\begin{tabular}{|p{15cm}|}
\hline
\textbf{Question}: (\entA{zoo lake}, located\_in\_the\_administrative\_territorial\_entity, ?) \\
\textbf{Answer}: \entD{gauteng} \\
\textbf{Rank-1 Path}: (zoo lake, Johannesburg, gauteng) \\
\underline{Passage1}: ... \entA{Zoo Lake} is a popular lake and public park in \entB{Johannesburg} , South Africa . It is part of the Hermann Eckstein Park and is ...\\
\underline{Passage2}: ... \entB{Johannesburg} ( also known as Jozi , Joburg and eGoli ) is the largest city in South Africa and is one of the 50 largest urban areas in the world . It is the provincial capital of \entD{Gauteng} , which is ... \\
\textbf{Rank-2 Path}: (zoo lake, South Africa, gauteng) \\
\underline{Passage1}: ... \entA{Zoo Lake} is a popular lake and public park in Johannesburg , \entB{South Africa} . It is ...\\
\underline{Passage2}: ... aka The Reef , is a 56-kilometre - long north - facing scarp in the \entD{Gauteng} Province of \entB{South Africa} . It consists of a ... \\
\hline
\hline
\textbf{Question}: (\entA{this day all gods die}, publisher, ?) \\
\textbf{Answer}: \entD{bantam books} \\
\textbf{Rank-1 Path}: (this day all gods die, Stephen R. Donaldson, bantam books) \\
\underline{Passage1}: ... All Gods Die , officially The Gap into Ruin : \entA{This Day All Gods Die} , is a science fiction novel by \entB{Stephen R. Donaldson} , being the final book of The Gap Cycle ...\\
\underline{Passage2}: ... The Gap Cycle ( published 19911996 by \entD{Bantam Books} and reprinted by Gollancz in 2008 ) is a science fiction story , told in a series of 5 books , written by \entB{Stephen R. Donaldson} . It is an ... \\
\textbf{Rank-2 Path}: (this day all gods die, Gap Cycle, bantam books) \\
\underline{Passage1}: ... All Gods Die , officially The Gap into Ruin : \entA{This Day All Gods Die} , is a science fiction novel by Stephen R. Donaldson , being the final book of The \entB{Gap Cycle} ...\\
\underline{Passage2}: ... The \entB{Gap Cycle} ( published 19911996 by \entD{Bantam Books} and reprinted by Gollancz in 2008 ) is a science fiction story ... \\
\hline
\end{tabular}
\caption{Two top-scoring paths for sample \wikihop Dev questions. In the Rank-1 path for the first question, the model composes the implicit \emph{located in} relations between (Zoo lake, Johannesburg) and (Johannesburg, Gauteng).}
\label{tab:extracted_paths}
\end{table*}

\subsection{Qualitative Analysis}
One key aspect of our model is its ability to indicate the paths that contribute most towards predicting an answer choice.
Table \ref{tab:extracted_paths} illustrates the two highest-scoring paths for two sample \wikihop questions which lead to correct answer prediction. In the first question, the top-2 paths are formed by connecting {\em Zoo Lake} to {\em Gauteng} through the intermediate entities {\em Johannesburg} and {\em South Africa}, respectively. 
In the second example, the science fiction novel {\em This Day All Gods Die} is connected to the publisher {\em Bantam Books} through the author {\em Stephen R.\ Donaldson}, and the collection {\em Gap Cycle} for first and second paths, respectively.

We also analyzed 50 randomly chosen questions that are annotated as requiring multi-hop reasoning in the \wikihop development set and that our model answered correctly. In 78\% of the questions, we found at least one meaningful path\footnote{A path is considered meaningful if it has valid relations that can be composed to conclude the predicted answer.} in the top-3 extracted paths, which dropped to 62\% for top-1 path. On average, 66\% of the top-3 paths returned by our model were meaningful. In contrast, only 46\% of three randomly selected paths per question made sense, even when limited to the paths for the correct answers. That is, a random baseline, even with oracle knowledge of the correct answer, would only find a good path in 46\% of the cases.
We also analyzed 50 questions that our model gets wrong. The top-scoring paths here were of lower quality (only 16.7\% were meaningful). This provides qualitative evidence that our model's performance is correlated with the quality of the paths it identifies, and it does not simply \emph{guess} using auxiliary information such as entity types, number of paths,\footnote{A model that returns the answer with the highest number of paths would score only 18.5\% on the \wikihop development set.} etc. 

\section{Conclusion}
\label{sec:conclusion}
We present a novel, path-based, multi-hop reading comprehension model that outperforms previous models on \wikihop and OpenBookQA.
Importantly, we illustrate how our model can explain its reasoning via explicit paths extracted across multiple passages. While we focused on 2-hop reasoning required by our evaluation datasets, the approach can be generalized to longer chains and to longer natural language questions.

\section*{Acknowledgment}
We thank Johannes Welbl for helping us evaluating our model on the WikiHop test set. We thank Rodney Kinney, Brandon Stilson and Tal Friedman for helping to produce the clean corpus used for OpenBookQA. We thank Dirk Groeneveld for retrieving the sentences from this corpus using the 2-hop retrieval. Computations on beaker.org were supported in part by credits from Google Cloud.

\bibliography{acl2019}
\bibliographystyle{acl_natbib}


\end{document}

%% file: macros.tex
\newcommand{\namecite}[1]{\citeauthor{#1}~\shortcite{#1}}

\newcommand{\colored}[2]{%
  \begingroup
  \setlength{\fboxsep}{1pt}%
  \colorbox{#1}{#2}%
  \endgroup
}
\newcommand{\entA}[1]{\colored{orange!50}{#1}}
\newcommand{\entB}[1]{\colored{olive!50}{#1}}
\newcommand{\entC}[1]{\colored{teal!50}{#1}}
\newcommand{\entD}[1]{\colored{green!50}{#1}}

\newcommand\T{\rule{0pt}{2.6ex}}       

\newcommand{\wikihop}{WikiHop\xspace}
\newcommand{\model}{PathNet: Path-based Multi-hop QA\xspace}
\newcommand{\modelname}{PathNet\xspace}

\newcommand{\psgs}{\ensuremath \mathcal{P}}
\newcommand{\psg}{\ensuremath \mathit{p}}
\newcommand{\pths}{\ensuremath \mathbb{P}}
\newcommand{\pth}{\ensuremath \mathsf{p}}

\newcommand{\ques}{\ensuremath \mathcal{Q}}

\newcommand{\psgsmat}{\ensuremath \mathbf{S}}

\newcommand{\locvec}{\ensuremath \mathbf{g}}

%% file: acl2019.bbl
\begin{thebibliography}{38}
\expandafter\ifx\csname natexlab\endcsname\relax\def\natexlab#1{#1}\fi

\bibitem[{Berant et~al.(2013)Berant, Chou, Frostig, and Liang}]{webquestions}
Jonathan Berant, Andrew Chou, Roy Frostig, and Percy Liang. 2013.
\newblock Semantic parsing on freebase from question-answer pairs.
\newblock In \emph{Proceedings of {EMNLP}}.

\bibitem[{Bollacker et~al.(2008)Bollacker, Evans, Paritosh, Sturge, and
  Taylor}]{bollacker2008freebase}
Kurt Bollacker, Colin Evans, Praveen Paritosh, Tim Sturge, and Jamie Taylor.
  2008.
\newblock Freebase: a collaboratively created graph database for structuring
  human knowledge.
\newblock In \emph{Proceedings of ACM SIGMOD international conference on
  Management of data}.

\bibitem[{Bordes et~al.(2015)Bordes, Usunier, Chopra, and
  Weston}]{simplequestions}
Antoine Bordes, Nicolas Usunier, Sumit Chopra, and Jason Weston. 2015.
\newblock Large-scale simple question answering with memory networks.
\newblock In \emph{NIPS}.

\bibitem[{Cao et~al.(2018)Cao, Aziz, and Titov}]{gcn_entity}
Nicola~De Cao, Wilker Aziz, and Ivan Titov. 2018.
\newblock Question answering by reasoning across documents with graph
  convolutional networks.
\newblock \emph{CoRR}, abs/1808.09920.

\bibitem[{Das et~al.(2017)Das, Neelakantan, Belanger, and
  McCallum}]{Das2017ChainsOR}
Rajarshi Das, Arvind Neelakantan, David Belanger, and Andrew McCallum. 2017.
\newblock Chains of reasoning over entities, relations, and text using
  recurrent neural networks.
\newblock In \emph{Proceedings of {EACL}}.

\bibitem[{Dhingra et~al.(2018)Dhingra, Jin, Yang, Cohen, and
  Salakhutdinov}]{corefgru}
Bhuwan Dhingra, Qiao Jin, Zhilin Yang, William Cohen, and Ruslan Salakhutdinov.
  2018.
\newblock Neural models for reasoning over multiple mentions using coreference.
\newblock In \emph{Proceedings of {NAACL}}.

\bibitem[{Dhingra et~al.(2017)Dhingra, Liu, Yang, Cohen, and
  Salakhutdinov}]{GAR}
Bhuwan Dhingra, Hanxiao Liu, Zhilin Yang, William Cohen, and Ruslan
  Salakhutdinov. 2017.
\newblock Gated-attention readers for text comprehension.
\newblock In \emph{Proceedings of {ACL}}.

\bibitem[{Fried et~al.(2015)Fried, Jansen, Hahn-Powell, Surdeanu, and
  Clark}]{fried2015higher}
Daniel Fried, Peter Jansen, Gustave Hahn-Powell, Mihai Surdeanu, and Peter
  Clark. 2015.
\newblock Higher-order lexical semantic models for non-factoid answer
  reranking.
\newblock \emph{{TACL}}, 3:197--210.

\bibitem[{Gardner and Mitchell(2015)}]{sfe}
Matt Gardner and Tom~M. Mitchell. 2015.
\newblock Efficient and expressive knowledge base completion using subgraph
  feature extraction.
\newblock In \emph{Proceedings of {EMNLP}}.

\bibitem[{Guu et~al.(2015)Guu, Miller, and Liang}]{kbpath}
Kelvin Guu, John Miller, and Percy Liang. 2015.
\newblock Traversing knowledge graphs in vector space.
\newblock In \emph{Proceedings of {EMNLP}}.

\bibitem[{Hochreiter and Schmidhuber(1997)}]{LSTM}
Sepp Hochreiter and J\"{u}rgen Schmidhuber. 1997.
\newblock Long short-term memory.
\newblock \emph{Neural Computation}, 9(8):1735--1780.

\bibitem[{Hu et~al.(2018)Hu, Peng, Huang, Qiu, Wei, and
  Zhou}]{hu2017reinforced}
Minghao Hu, Yuxing Peng, Zhen Huang, Xipeng Qiu, Furu Wei, and Ming Zhou. 2018.
\newblock Reinforced mnemonic reader for machine reading comprehension.
\newblock In \emph{Proceedings of {IJCAI}}.

\bibitem[{Jansen et~al.(2017)Jansen, Sharp, Surdeanu, and
  Clark}]{Jansen2017FramingQA}
Peter Jansen, Rebecca Sharp, Mihai Surdeanu, and Peter Clark. 2017.
\newblock Framing qa as building and ranking intersentence answer
  justifications.
\newblock \emph{Computational Linguistics}, 43:407--449.

\bibitem[{Joshi et~al.(2017)Joshi, Choi, Weld, and
  Zettlemoyer}]{joshi-EtAl:2017:Trivia-qa}
Mandar Joshi, Eunsol Choi, Daniel Weld, and Luke Zettlemoyer. 2017.
\newblock {TriviaQA}: {A} large scale distantly supervised challenge dataset
  for reading comprehension.
\newblock In \emph{Proceedings of {ACL}}.

\bibitem[{Khashabi et~al.(2019)Khashabi, Azer, Khot, Sabharwal, and
  Roth}]{Khashabi2019OnTC}
Daniel Khashabi, Erfan~Sadeqi Azer, Tushar Khot, Ashutosh Sabharwal, and Dan
  Roth. 2019.
\newblock On the capabilities and limitations of reasoning for natural language
  understanding.
\newblock \emph{CoRR}, abs/1901.02522.

\bibitem[{Khashabi et~al.(2018{\natexlab{a}})Khashabi, Chaturvedi, Roth,
  Upadhyay, and Roth}]{MultiRCKhashabi2018}
Daniel Khashabi, Snigdha Chaturvedi, Michael Roth, Shyam Upadhyay, and Dan
  Roth. 2018{\natexlab{a}}.
\newblock Looking beyond the surface: A challenge set for reading comprehension
  over multiple sentences.
\newblock In \emph{Proceedings of {NAACL}}.

\bibitem[{Khashabi et~al.(2016)Khashabi, Khot, Sabharwal, Clark, Etzioni, and
  Roth}]{tableilp2016}
Daniel Khashabi, Tushar Khot, Ashish Sabharwal, Peter Clark, Oren Etzioni, and
  Dan Roth. 2016.
\newblock Question answering via integer programming over semi-structured
  knowledge.
\newblock In \emph{Proceedings of {IJCAI}}.

\bibitem[{Khashabi et~al.(2018{\natexlab{b}})Khashabi, Khot, Sabharwal, and
  Roth}]{semanticilp2018aaai}
Daniel Khashabi, Tushar Khot, Ashish Sabharwal, and Dan Roth.
  2018{\natexlab{b}}.
\newblock Question answering as global reasoning over semantic abstractions.
\newblock In \emph{Proceedings of {AAAI}}.

\bibitem[{Khot et~al.(2017)Khot, Sabharwal, and Clark}]{Khot2017AnsweringCQ}
Tushar Khot, Ashish Sabharwal, and Peter Clark. 2017.
\newblock Answering complex questions using open information extraction.
\newblock In \emph{Proceedings of {ACL}}.

\bibitem[{Kingma and Ba(2015)}]{adam}
Diederik~P. Kingma and Jimmy~Lei Ba. 2015.
\newblock Adam: {A} method for stochastic optimization.
\newblock In \emph{Proceedings of {ICLR}}.

\bibitem[{Lai et~al.(2017)Lai, Xie, Liu, Yang, and Hovy}]{Lai2017RACELR}
Guokun Lai, Qizhe Xie, Hanxiao Liu, Yiming Yang, and Eduard~H. Hovy. 2017.
\newblock Race: Large-scale reading comprehension dataset from examinations.
\newblock In \emph{EMNLP}.

\bibitem[{Lao et~al.(2011)Lao, Mitchell, and Cohen}]{pra}
Ni~Lao, Tom~M. Mitchell, and William~W. Cohen. 2011.
\newblock Random walk inference and learning in a large scale knowledge base.
\newblock In \emph{Proceedings of {EMNLP}}.

\bibitem[{Mihaylov et~al.(2018)Mihaylov, Clark, Khot, and
  Sabharwal}]{openbookqa}
Todor Mihaylov, Peter Clark, Tushar Khot, and Ashish Sabharwal. 2018.
\newblock Can a suit of armor conduct electricity? {A} new dataset for open
  book question answering.
\newblock In \emph{Proceedings of {EMNLP}}.

\bibitem[{Pennington et~al.(2014)Pennington, Socher, and
  Manning}]{pennington2014glove}
Jeffrey Pennington, Richard Socher, and Christopher~D. Manning. 2014.
\newblock {GloVe}: Global vectors for word representation.
\newblock In \emph{Proceedings of {EMNLP}}.

\bibitem[{Peters et~al.(2018)Peters, Neumann, Iyyer, Gardner, Clark, Lee, and
  Zettlemoyer}]{elmo}
Matthew~E. Peters, Mark Neumann, Mohit Iyyer, Matt Gardner, Christopher Clark,
  Kenton Lee, and Luke Zettlemoyer. 2018.
\newblock Deep contextualized word representations.
\newblock In \emph{Proceedings of {NAACL}}.

\bibitem[{Radford et~al.(2018)Radford, Narasimhan, Salimans, and
  Sutskever}]{radford2018improving}
Alec Radford, Karthik Narasimhan, Tim Salimans, and Ilya Sutskever. 2018.
\newblock Improving language understanding by generative pre-training.
\newblock Technical report, Technical report, OpenAI.

\bibitem[{Rajpurkar et~al.(2016)Rajpurkar, Zhang, Lopyrev, and
  Liang}]{Rajpurkar2016-squad}
Pranav Rajpurkar, Jian Zhang, Konstantin Lopyrev, and Percy Liang. 2016.
\newblock {SQuAD}: 100,000+ questions for machine comprehension of text.
\newblock In \emph{Proceedings of {EMNLP}}.

\bibitem[{Seo et~al.(2017)Seo, Kembhavi, Farhadi, and Hajishirzi}]{bidaf}
Minjoon Seo, Aniruddha Kembhavi, Ali Farhadi, and Hannaneh Hajishirzi. 2017.
\newblock Bidirectional attention flow for machine comprehension.
\newblock In \emph{Proceedings of {ICLR}}.

\bibitem[{Shen et~al.(2017)Shen, Huang, Gao, and Chen}]{shen2017reasonet}
Yelong Shen, Po-Sen Huang, Jianfeng Gao, and Weizhu Chen. 2017.
\newblock Reasonet: Learning to stop reading in machine comprehension.
\newblock In \emph{Proceedings of {KDD}}.

\bibitem[{Song et~al.(2018)Song, Wang, Yu, Zhang, Florian, and
  Gildea}]{mhqa_grn}
Linfeng Song, Zhiguo Wang, Mo~Yu, Yue Zhang, Radu Florian, and Daniel Gildea.
  2018.
\newblock Exploring graph-structured passage representation for multi-hop
  reading comprehension with graph neural networks.
\newblock \emph{CoRR}, abs/1809.02040.

\bibitem[{Srivastava et~al.(2014)Srivastava, Hinton, Krizhevsky, Sutskever, and
  Salakhutdinov}]{dropout}
Nitish Srivastava, Geoffrey Hinton, Alex Krizhevsky, Ilya Sutskever, and Ruslan
  Salakhutdinov. 2014.
\newblock Dropout: A simple way to prevent neural networks from overfitting.
\newblock \emph{{JMLR}}, 15(1):1929--1958.

\bibitem[{Sun et~al.(2018)Sun, Yu, Yu, and Cardie}]{Sun2018ImprovingMR}
Kai Sun, Dian Yu, Dong Yu, and Claire Cardie. 2018.
\newblock Improving machine reading comprehension with general reading
  strategies.
\newblock \emph{CoRR}, abs/1810.13441.

\bibitem[{Toutanova et~al.(2016)Toutanova, Lin, tau Yih, Poon, and
  Quirk}]{Toutanova2016CompositionalLO}
Kristina Toutanova, Victoria Lin, Wen tau Yih, Hoifung Poon, and Chris Quirk.
  2016.
\newblock Compositional learning of embeddings for relation paths in knowledge
  base and text.
\newblock In \emph{Proceedings of {ACL}}.

\bibitem[{Trischler et~al.(2017)Trischler, Wang, Yuan, Harris, Sordoni,
  Bachman, and Suleman}]{Trischler2017-rc-newsqa}
Adam Trischler, Tong Wang, Xingdi Yuan, Justin Harris, Alessandro Sordoni,
  Philip Bachman, and Kaheer Suleman. 2017.
\newblock {NewsQA}: {A} machine comprehension dataset.
\newblock In \emph{Proceedings of the 2nd Workshop on Representation Learning
  for NLP}.

\bibitem[{Welbl et~al.(2018)Welbl, Stenetorp, and Riedel}]{wikihop}
Johannes Welbl, Pontus Stenetorp, and Sebastian Riedel. 2018.
\newblock Constructing datasets for multi-hop reading comprehension across
  documents.
\newblock \emph{{TACL}}, 6:287--302.

\bibitem[{Weston et~al.(2015)Weston, Bordes, Chopra, and
  Mikolov}]{babi-Weston-15}
Jason Weston, Antoine Bordes, Sumit Chopra, and Tomas Mikolov. 2015.
\newblock Towards {AI}-complete question answering: {A} set of prerequisite toy
  tasks.
\newblock \emph{CoRR}, abs/1502.05698.

\bibitem[{Wilson(1927)}]{wilson}
Edwin~B. Wilson. 1927.
\newblock Probable inference, the law of succession, and statistical inference.
\newblock \emph{JASA}, 22(158):209--212.

\bibitem[{Zhong et~al.(2019)Zhong, Xiong, Keskar, and Socher}]{cfc}
Victor Zhong, Caiming Xiong, Nitish~Shirish Keskar, and Richard Socher. 2019.
\newblock Coarse-grain fine-grain coattention network for multi-evidence
  question answering.
\newblock In \emph{Proceedings of {ICLR}}.

\end{thebibliography}
